\documentclass[runningheads]{llncs}

\usepackage[T1]{fontenc}
\usepackage{multirow}
\usepackage[title]{appendix}
\usepackage{subfig}
\usepackage{graphicx}
\usepackage{wrapfig}
\usepackage{blindtext}
\usepackage{amsmath}
\usepackage{amssymb}
\usepackage{booktabs}
\usepackage{siunitx}
\usepackage{pifont}
\usepackage[misc,geometry]{ifsym}
\usepackage[hidelinks,colorlinks=true,allcolors=blue]{hyperref}
\newcommand{\etal}{\textit{et al.}}
\newcommand{\cmark}{\ding{51}}%
\newcommand{\xmark}{\ding{55}}%
\newcommand{\pubtables}{PubTables-1M~\cite{pubtables}}
\usepackage{color}

\makeatletter
\renewcommand\subsubsection{\@startsection{subsubsection}{3}{\z@}%
	{-8\p@ \@plus -4\p@ \@minus -4\p@}%
	{-0.5em \@plus -0.22em \@minus -0.1em}%
	{\normalfont\normalsize\bfseries\boldmath}}
\makeatother

\begin{document}
	\title{Latent Diffusion for Guided Document Table Generation}
	\author{
		Syed Jawwad Haider Hamdani\inst{1,2,3}\orcidID{0000-1111-2222-3333} \and
		Saifullah Saifullah\inst{1,2} \orcidID{0000-0003-3098-2458} \and
		Stefan Agne\inst{1,3} \orcidID{0000-0002-9697-4285} \and
		Andreas Dengel\inst{1,2} \orcidID{0000-0002-6100-8255} \and
		Sheraz Ahmed\inst{1,3} \orcidID{0000-0002-4239-6520}}
	\authorrunning{Syed Jawwad Haider Hamdani \textit{et al.}}
	\institute{
		Smarte Daten and Wissensdienste (SDS), Deutsches Forschungszentrum für Künstliche Intelligenz GmbH (DFKI), Trippstadter Straße 122,
		67663 Kaiserslautern, Germany\\\email{\{firstname.lastname\}@dfki.de}\\ \and
		Department of Computer Science, RPTU Kaiserslautern-Landau, Erwin-Schrödinger-Straße 52, 67663 Kaiserslautern, Germany\and
		DeepReader GmbH, 67663 Kaiserlautern, Germany\\
	}
	\maketitle              %
	\begin{abstract}
		
		Obtaining annotated table structure data for complex tables is a challenging task due to the inherent diversity and complexity of real-world document layouts. The scarcity of publicly available datasets with comprehensive annotations for intricate table structures hinders the development and evaluation of models designed for such scenarios. This research paper introduces a novel approach for generating annotated images for table structure by leveraging conditioned mask images of rows and columns through the application of latent diffusion models. The proposed method aims to enhance the quality of synthetic data used for training object detection models. Specifically, the study employs a conditioning mechanism to guide the generation of complex document table images, ensuring a realistic representation of table layouts. To evaluate the effectiveness of the generated data, we employ the popular YOLOv5 object detection model for training. The generated table images serve as valuable training samples, enriching the dataset with diverse table structures. The model is subsequently tested on the challenging pubtables-1m testset, a benchmark for table structure recognition in complex document layouts. Experimental results demonstrate that the introduced approach significantly improves the quality of synthetic data for training, leading to YOLOv5 models with enhanced performance. The mean Average Precision (mAP) values obtained on the pubtables-1m testset showcase results closely aligned with state-of-the-art methods. Furthermore, low FID results obtained on the synthetic data further validate the efficacy of the proposed methodology in generating annotated images for table structure.
	\end{abstract}
	\section{Introduction}
    The task of table structure recognition (TSR) holds significant importance in the field of document analysis~\cite{tsrformer,layout-shen2021,tableformer,tablebank,pubtables}, as tables frequently appear in business documents to summarize critical information~\cite{ICDAR-2013,ICDAR-2017,tablebank}.
	However, the abundance of highly diverse and unstructured layouts of tables in the document data~\cite{ICDAR-2013,ICDAR-2017,pubtabnet,fintabnet,scitsr,tablebank,pubtables} makes it a challenging task. Despite the great success of data-driven deep learning (DL)-based approaches on TSR in recent years ~\cite{tsrformer,layout-shen2021,tableformer}, it still remains difficult to apply these approaches to private real-world datasets, primarily attributed to two main reasons: costs associated with table structure annotation and large distribution shift between public and private datasets. 	
	
	While a wide range of publicly-available table recognition datasets have been proposed in the recent years~\cite{ICDAR-2013,ICDAR-2017,pubtabnet,fintabnet,scitsr,tablebank,pubtables} to improve recent DL-based approaches, they are either manually annotated~\cite{ICDAR-2013,ICDAR-2017} or generated through automated parsing of metadata of HTML or PDF-based documents~\cite{tablebank,pubtables}. 
	Since precisely annotating table datasets for structure recognition can be significantly labor-intensive and time-consuming, resulting in huge annotation costs for large and diverse datasets, the majority of publicly available datasets with manual annotations are small-scale~\cite{ICDAR-2013,ICDAR-2017}.
	On the other hand, datasets that are automatically generated are large-scale and incur little annotation costs, but are often accompanied by significant annotation noise~\cite{tables-survey}. 
	Another issue is that the distributions of these datasets typically vary significantly compared to real enterprise datasets. 
	In particular, the large-scale automatically parsed datasets are typically generated using publicly available online (academic publication or government procedural) documents and therefore show very low variance in the data~\cite{publaynet,pubtables}, hardly reflecting the real-world complexity of printed documents, which often appear under various forms of noise or corruption~\cite{saifullah-robustness,ShabbyPages2023}. 
	This limited sample diversity in these datasets enables DL-based approaches to achieve promising results; however, achieving the same performance on more complex and diverse real-world datasets remains difficult.
	
	In this paper, we aim to tackle the aforementioned challenges of table annotation and sample diversity through data synthesis. 	
	In particular, we approach the task of generating table datasets through guided image synthesis, leveraging the recently introduced diffusion models (DMs)~\cite{ddpm}.	
	DMs~\cite{ddpm} have recently demonstrated exceptional performance in natural image synthesis~\cite{ddpm,ldms,saharia2022palette}, surpassing generative adversarial networks (GANs)~\cite{gans} in both fidelity and sample diversity, and have gained considerable attention in the domain of document analysis~\cite{diff-image-gen,diff-layout-1,diff-layout-2,diff-binarization,diff-enhancement}. 
	This can be primarily attributed to their unique properties, such as the ability to avoid mode-collapse and training instabilities commonly observed in GANs~\cite{thanhtung2020catastrophic}, and the ease of introducing conditional generation in these models~\cite{saharia2022palette,ldms}. 	
	In this work, we specifically explore latent diffusion models (LDMs)~\cite{ldms} for the task of document table synthesis and propose a conditional LDM that conditions the generation process on an input mask that describes the structure of the table. 
	Consequently, simply by varying the conditioning mask, a variety of tables can be generated. Overall, the contributions of this paper can be summarized as follows:
	\begin{itemize}
		\item Generating synthetic document table images using row/column conditioning masks in latent diffusion models 
		\item Ensuring the viability of the generated synthetic data by achieving low FID scores.
		\item Obtaining close to benchmark performances for table structure recognition tasks by augmenting real-world datasets with our synthetic data. 
	\end{itemize}

    \section{Related Work}
    \subsection{Generative Models in Image Synthesis}
    The task of image synthesis is of great importance in computer vision and has been extensively explored in the past decade~\cite{gans,dcgan,stylegan,dalle,vqvae,saharia2022palette,ldms,diffTransformer}. 
    The most popular approaches in this domain include the previous state-of-the-art (SotA), including variational autoencoders (VAEs)~\cite{vaes}, autoregressive models~\cite{pixelcnn,dalle,vqvae}, and generative adversarial networks (GANs)~\cite{gans,stylegan,dcgan}, as well as the most recent SotA, diffusion models (DMs)~\cite{ddpm,ldms,imagen,dalle2}, each exhibiting its own set of advantages and disadvantages.
    For instance, GANs~\cite{gans,stylegan,dcgan} are renowned for their capability to generate high-resolution images; however, they are also notoriously challenging to train, often facing mode collapse during the training process. 	
    VAEs~\cite{vaes,vqvae}, on the other hand, are easier to train and offer faster image generation times; however, their image quality and sample diversity is not as impressive compared to GANs~\cite{gans,stylegan,dcgan} or DMs~\cite{ddpm,ldms,imagen}. 	
    Autoregressive models~\cite{pixelcnn} are also well-known for their performance in generation tasks, but they suffer from the issue of extremely slow sampling, resulting in huge inference costs, especially when working with high-resolution images.	
    
    Diffusion models (DMs)~\cite{ddpm,ldms,imagen,dalle2,diffTransformer} have recently outperformed many of the existing state-of-the-art (SotA) generative approaches such as StyleGAN3~\cite{stylegan} and DALLE~\cite{dalle}, in terms of both image quality and sample diversity, demonstrating exceptional performance on a wide variety of natural image benchmarks~\cite{ddpm,ldms,imagen}. DMs~\cite{ddpm,ldms,imagen,dalle2} also present several advantages over GANs~\cite{gans,stylegan}, such as training stability and ease of introducing conditions into the training process. 
    However, the originally proposed diffusion models~\cite{ddpm} also suffer from extremely high sampling times, especially when generating high-resolution images.
    To address this issue, Rombach et al.~\cite{ldms} recently introduced latent diffusion models (LDMs) that compress the images to low-dimensional latent representations using an autoencoder and train a diffusion model to generate images directly in the latent space, effectively reducing the high inference costs incurred by the sampling process.
    
    \subsection{Diffusion Models in Document AI}
    Diffusion models (DMs)~\cite{ddpm,ldms} have been making waves in the document AI community, with several recent works proposing DMs for a diverse range of document data generation tasks~\cite{diff-binarization,diff-enhancement,diff-image-gen,diff-layout-1,diff-layout-2,diffusion-based-text-gen}. 
    Inoue~\etal~\cite{diff-layout-1} recently extended discrete diffusion models (D3PMs)~\cite{d3pm} for controlled generation of document layouts. Similarly, He~\etal~\cite{diff-layout-2} investigated document layout generation using sequential diffusion, generating layouts as a sequence of structural elements. 		
    In a different direction, Saifullah~\etal~\cite{diff-binarization} and Zongyuan~\etal~\cite{diff-enhancement} explored image-to-image diffusion models for the tasks of document image binarization enhancement. 			
    In a number of recent works, handwritten text generation~\cite{diffusion-based-text-gen,diffusion-based-text-gen-2,luhman2020diffusion} and scene text generation~\cite{diffusion-based-text-gen-2} have also been investigated using text-to-image diffusion models~\cite{ldms}.	
    The most relevant to our work is the recent study by Tanveer~\etal~\cite{diff-image-gen}, which introduced a layout-guided diffusion model for generating full-page document images. Their results, however, were limited to low to medium resolutions, and the layout conditioning mechanism relied on learning separate bounding box embeddings for each layout element using a separate Transformer-based model~\cite{transformers}. 
    
    \section{Diffusion for Guided Table Generation}
    \begin{figure}[t!]
    \begin{center}
      \includegraphics[width=\textwidth]{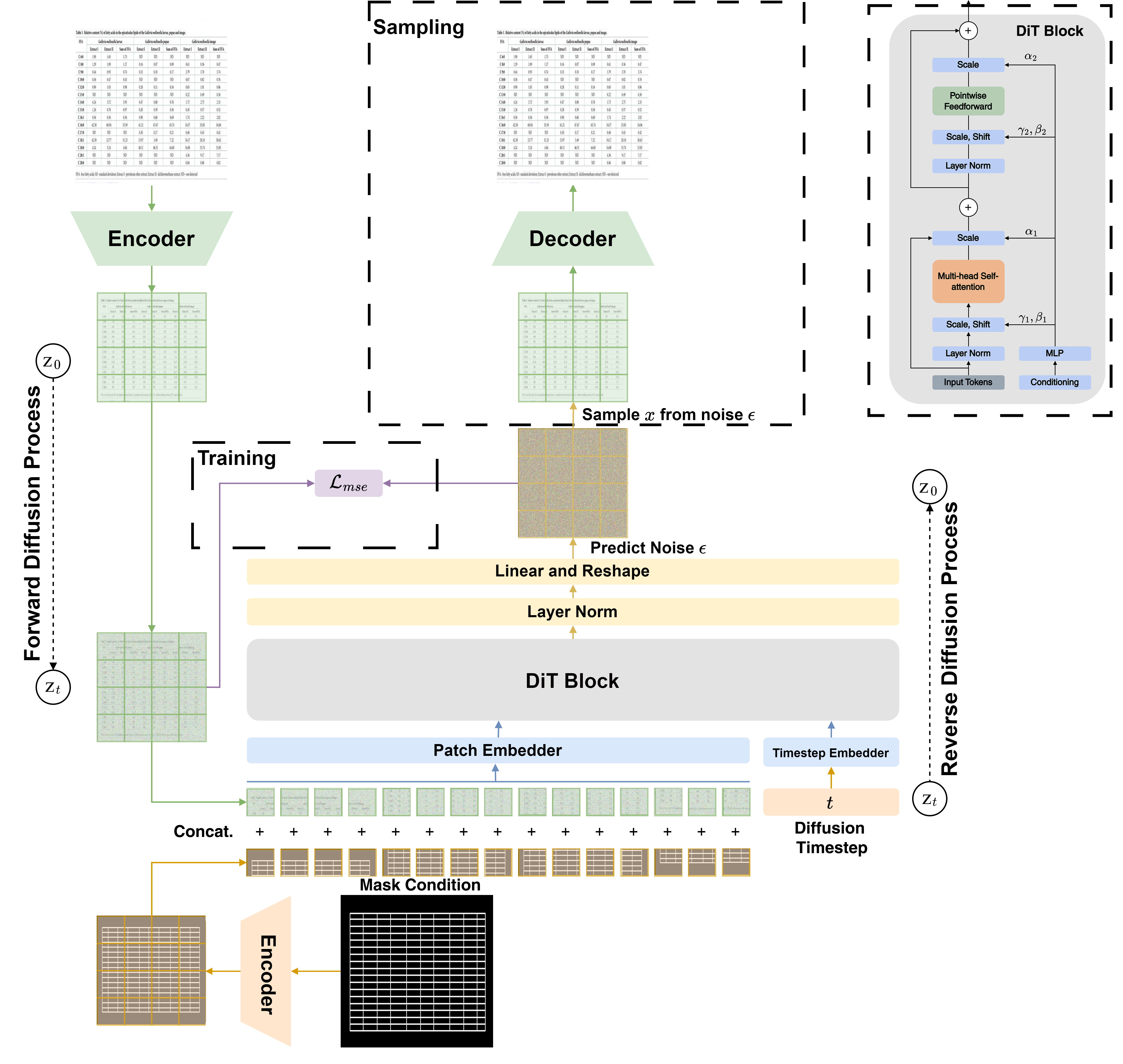}
    \end{center}
    \caption{The entire model depicting our methodology. $z_0$ is the latent representation of input document table image $x$, $z_t$ is the is the noisy latent after forward diffusion, Mask Condition is the conditioning mask image that depicts the row/column structure of the training table image, Sampling region depicts the newly generated synthetic image.} 
    \label{fig:approach}
    \end{figure}
	In this section, we present our proposed approach for guided document table synthesis, the entire workflow of which is illustrated in Fig.~\ref{fig:approach}. 
	Formally, let $\mathbb{D}=\{(x_1, y_1),(x_2, y_2),\dots,(x_N, y_N) \}$ define a dataset comprised of pairs document table images $x \in \mathbb{R}^{C\times H\times W}$ and their corresponding ground-truth table structure annotations $y \in \mathbb{Y}$. 
	First, we utilize an autoencoder~\cite{vaes} model to encode the input document image table $x \in \mathbb{R}^{C\times H\times W}$ into a compressed latent representation $z \in \mathbb{R}^{\hat{C}\times \hat{H}\times \hat{W}}$, a step that is necessary to reduce both the training and sampling costs of the generation process, especially when working on large-resolution images. Then, a diffusion model is trained end-to-end in a conditional manner, where the condition is defined by a binary mask that represents the target structure of the synthesized table. Finally, in order to generate new table images, a conditional sampling process followed by a autoencoder decoding step is utilized.  
	In the following sections, we provide a more detailed explanation of the latent space compression, the forward and conditional reverse diffusion processes employed in our approach.

    \subsection{Autoencoder Definition}    
    \label{sec:autoencoder}
    To compress the input table images into latent space representation, we utilize a publicly available variational autoencoder (VAE) model that is originally trained on the large-scale natural image dataset LAION-5B~\cite{schuhmann2022laionb}. During our preliminary experiments, we found the model reconstruction accuracy of the model to be sufficient therefore we did not find the need to further finetune it on the document distribution.
    Given an input document table image $x \in \mathbb{R}^{3 \times H \times W}$ in the RGB space, the encoder $\mathcal{E}$ encodes $x$ into a latent representation $z = \mathcal{E}(x)$, where $z \in \mathbb{R}^{4\times \frac{H}{f}\times \frac{W}{f}}$, where the compression factor of the encoder is $f=8$, greatly reducing the size at which images are processed during both training and sampling phase. Conversely, the decoder $\mathcal{D}$ reconstructs the image from a given latent representation to the image space $\tilde{x} = \mathcal{D}(z)$.

    \subsection{Forward Diffusion}    
    To train the diffusion model for table image synthesis, we use the standard forward diffusion process~\cite{ddpm} which involves taking an input image in the latent space $z \in \mathbb{R}^{4\times \frac{H}{f}\times \frac{W}{f}}$ and gradually injecting it with noise to transform it into a corresponding noisy latent image $z_t$. Typically, forward diffusion is defined as a fixed Markov process that systematically introduces Gaussian noise $\sim \mathcal{N}$ to the latent image $z_0$ over a total of $T$ iterations as follows:
    \begin{align}
        q(z_1, \ldots, z_T | z_0) &:= \prod_{t=1}^{T} q(z_t | z_{t-1})\\
        q(z_t | z_{t-1}) &:= \mathcal{N}(z_t; \sqrt{1-\beta_t} z_{t-1}, \beta_t\textbf{I})
    \end{align}
    where $\beta_1,\dots,\beta_T$ are the hyperparameters that define the variance of the noise added at each timestep $t$ and are subject to the constraint $0<\beta_t<1$. $z_t$ is the output noisy latent image at each timestep $t$.  Importantly, one can express the above forward chain directly in terms of the $z_0$ as follows:
    \begin{align}
     q(z_t | z_0) &:= \mathcal{N}(z_t; \sqrt{\hat{\alpha_t}} z_{t-1}, (1-\hat{\alpha_t})\textbf{I})\\
     \hat{\alpha} &:= \prod_{s=1}^{t}\alpha_t, \alpha_t = 1-\beta_t
    \end{align}
    This allows expressing the noisy latent image $z_t$ at any step $t$ directly in terms of the noise $\epsilon \sim \mathcal{N}(\textbf{0}, \textbf{I})$:
    \begin{equation}
    	z_t = \sqrt{\hat{\alpha_t}}z_0 + (1-\hat{\alpha_t}) \epsilon
    \end{equation}
	This forward diffusion process can be visualized in Fig.~\ref{fig:approach} where the input latent image $z_0$ is diffused into an intermediate noisy representation $z_t$ during the training process.

    \begin{figure}[b!]
    \centering
    \includegraphics[width=\textwidth]{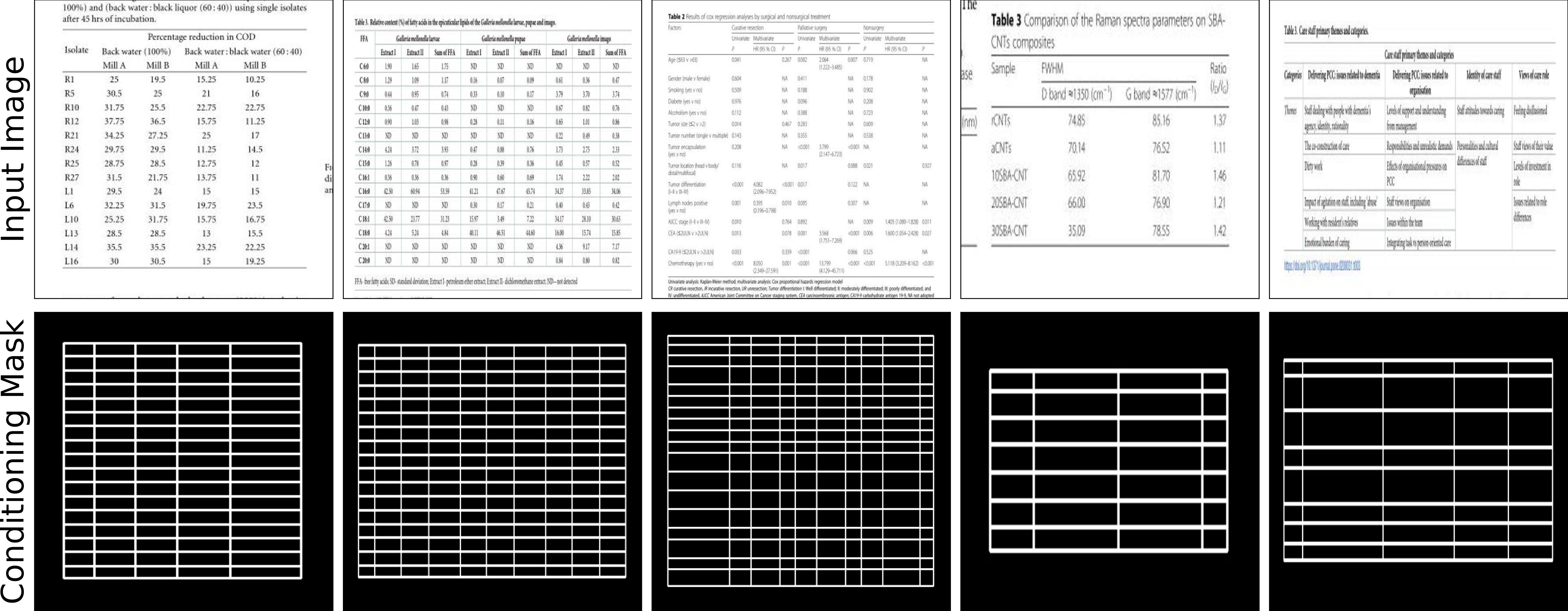}
    \hfill
    \caption{A few sample images from the \pubtables{} dataset (top), along with their corresponding mask images depicting row and column structures (bottom), are shown.}
    \label{fig:mask_condition}
    \end{figure}

    \subsection{Reverse Diffusion}
	The reverse diffusion process is expressed as a inverse Markov chain $p_\theta(z_0, \ldots, z_{t-1}|z_T)$ that,  starting at $p(z_T) \sim \mathcal{N}(\textbf{0}, \textbf{I})$, reverts the forward noising process to gradually restore the original latent image $z_0$ back from the noisy image $z_t$ with learned transition steps $p_\theta(z_{t-1}|z_{t})$. 
	Formally, in an unconditional setting, the reverse diffusion process can be expressed as follows:
	\begin{align} 
		p(z_{0:T}) &:= p(z_T)\prod_{t=1}^{T} p_\theta(z_{t-1}|z_t)\\
		p_\theta(z_{t-1}|z_t) &:= \mathcal{N}(z_{t-1}; \mu_\theta(z_t, t), \Sigma_\theta (z_t, t))
	\end{align}
	In this work, to guide the generation process according to the requirement of the table structure, we introduce additional condition into the generation process. Recall that the original ground-truth annotation $y$ for each input table image $x \in R^{C\times H\times W}$ in the original dataset may be present in any given format, such as, in JSON or XML file formats as lists of bounding boxes. Then, let $g: \mathbb{Y}  \rightarrow \{0, 1\}^{H\times W}$ be any function that maps a given table structure annotation $y \in \mathbb{Y}$ into a corresponding binary mask $m\in \{0, 1\}^{H\times W}$ where each row and column of the original annotation is set to 1 whereas the rest of the image is set to 0 as shown in Fig.~\ref{fig:approach}. A few example images of such masks that are generated from the corresponding table annotations are shown in Fig.~\ref{fig:mask_condition}. After generating the corresponding annotation masks $m=g(y)$ for each input input annotation $y$, we introduce them into the model as an additional condition to be used for each learned $p_\theta(z_{t-1}|z_{t}, m)$, resulting in the following reverse process:
	\begin{align} 
		p(z_{0:T}|m) &:= p(z_T)\prod_{t=1}^{T} p_\theta(z_{t-1}|z_t,m)\\
		p_\theta(z_{t-1}|z_t,m) &:= \mathcal{N}(z_{t-1}; \mu_\theta(z_t, t,m), \Sigma_\theta (z_t, t,m))
	\end{align}
	where the learned transition $p_\theta(z_{t-1}|z_t,m)$ is implemented in terms of a diffusion transformer network~\cite{diffTransformer} $f_\theta$ (see Fig.~\ref{fig:approach}) which is trained to predict the noise introduced $\epsilon\sim \mathcal{N}(\textbf{0}, \textbf{I})$, allowing to approximate the input latent image as follows:
	\begin{equation}
		\hat{z}_0 = \frac{1}{\sqrt{\hat{\alpha_t}}} (z_t - \sqrt{1-\hat{\alpha_t}}f_\theta(z_t, m, c, t))
	\end{equation}
	Finally, we train the noise prediction network $f_\theta$ by minimizing the following loss:
	\begin{equation}
	\mathcal{L}_{mse} = \mathbb{E}_{z, m, \epsilon \sim \mathcal{N}(0, 1), t} \left[\lvert\lvert\epsilon -  f_{\theta}(z_t, m, c, t) \rvert\rvert_2^2 \right]
	\end{equation}

    \subsection{Diffusion Transformer}
    We define the noise prediction network $f_{\theta}(z_t, t, m)$ as a diffusion transformer~\cite{diffTransformer} with 12 layers, each with an embedding dimension of size 768. In each DiT block, we utilize adaptive layer normalization (AdaLN-norm) which was demonstrated to perform the best on natural image generation tasks~\cite{diffTransformer}. Both the latent representations of the input noisy images $x$ and the corresponding mask images $m$ are first divided into patches of size $p\times p$ and then concatenated across the channel dimension. This enables directly conditioning the noisy image onto the additional guidance information received from the annotation masks $m$.  A detailed view of the diffusion transformer block can be visualized in Fig.~\ref{fig:approach}.

    \subsection{Sampling}
    To generate new table images after the training process, we utilize the standard DDIM sampler proposed in~\cite{song2021denoising}, which starting from complete noise $z_T\sim \mathcal{N}(\textbf{0}, \textbf{I})$, where $T=1000$, iteratively applies the reverse diffusion step $p_\theta(z_{t-1}|t,m)$ to generate a new image sample.	In all our experiments, we use a total of 750 intermediate sampling steps to generate the samples.
        
	\section{Experimental Setup}
	In this section, we outline the configuration of our experiments, which includes details about the datasets utilized, evaluation metrics, and the training process. 	
 	\subsubsection{Datasets.} 
	To train the diffusion models for our task, we utilize the large-scale \pubtables{} dataset, which is specifically designed for table detection and structure recognition tasks. 
	The dataset contains 575,305 annotated document pages containing tables for the table detection tasks. Additionally, it includes 947,642 fully annotated tables with complete location (bounding box) information for table structure recognition and functional analysis tasks.	
	Since the primary focus of this paper is structure-conditioned table generation, we utilize the 947,642 document table images available in the dataset which are split into training, testing and validation splits of sizes 758,849, 93,834, and 94,959 respectively. The annotations for these tables are originally present in the XML format with bounding box information provided in PASCAL VOC format. 
	We also utilize another dataset solely for evaluation purposes, namely the small-scale ICDAR2013 competition table recognition dataset. We were only able to find the validation and test-set for this dataset which comprised of 100 and 158 images respectively.

	\begin{table}[b!]
		\centering
		\caption{Training configurations for our experiments with \pubtables{} dataset.}\label{tab:exp_setup}
	\resizebox{\textwidth}{!}{
		\setlength{\tabcolsep}{0.5em} %
		\begin{tabular}{ccccc}
			\toprule
			Image Dimension & Latent Dimension & Mask Conditioning & No. of samples & Training iterations\\
			\midrule
			$256\times 256\times 3$ & $32\times 32\times 4$  & \xmark & $450$k & $350$k \\
			$512\times 512\times 3$ & $64\times 64\times 4$ & \xmark & $450$k & $350$k \\
			$256\times 256\times 3$ & $32\times 32\times 4$ & \cmark & $450$k & $450$k \\
			$512\times 512\times 3$ & $64\times 64\times 4$  & \cmark & $450$k & $450$k \\
			\bottomrule
		\end{tabular}
	}
	\end{table} 
	\subsubsection{Data Preparation.} 
	To generate the structure conditioning masks for training, we rely on the row and column annotations found in the training set. We start by creating a blank mask image $m$ for each corresponding train image $x$, and then proceed to utilize the ground-truth annotations to draw the rows on columns on the mask $m$. Afterwards, both table images and their corresponding masks are converted to RGB space, normalized and then fed into the pretrained VAE (as mentioned in Section.~\ref{sec:autoencoder}) to prepare their corresponding latent space representations. The pretrained weights which were used to initialize the VAE are provided by StabilityAI~\footnote{\url{https://huggingface.co/stabilityai/sd-vae-ft-ema}}. Since we experiment with two different resolutions in this work, $256\times 256$ and $512\times 512$, their corresponding latent space representations turn out to be of sizes $32\times 32\times 4$ and $64\times 64\times 4$, respectively. After encoding, all latent space representations are normalized with a fixed scaling factor to have a unit variance. It is worth highlighting that since the VAE is not trained as part of the process and only used with frozen weights to generate the latent space representations, we compute all the latents for the target datasets only once and store them on disk during training, allowing us to efficiently perform multiple experiments.

	\subsubsection{Training Setup.} 
	As outlined in Table~\ref{tab:exp_setup}, we train our table generation models for two target resolutions, $256\times 256$ and $512\times 512$, and in addition to our proposed structure-conditional setting, we also train the models under unconditional setting for comparison.
	All models were trained on 450k samples from the \pubtables{} dataset, undergoing 350k training steps for the unconditional setting and 450k training steps for the proposed structure-conditioned setting.
	In addition, a distributed training setup was employed with 4-8 NVIDIA A100 GPUs, whereas the optimization was performed using the Adam optimizer with a learning rate of 1.0e-4.
	
    \section{Results}
    
    \begin{figure}[h!]
    \begin{center}
      \includegraphics[width=\textwidth]{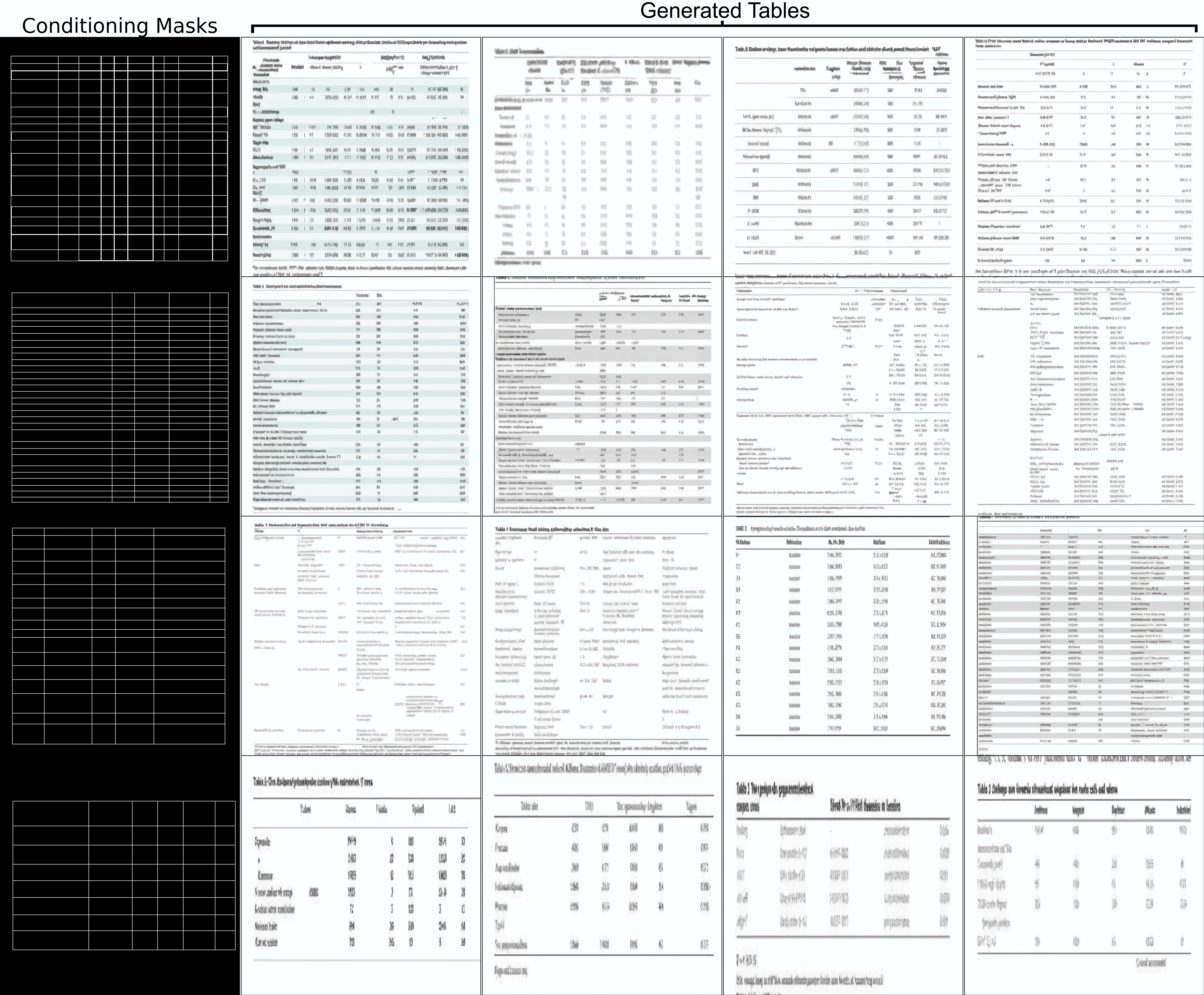}
    \end{center}
    \caption{Multiple image generations from single mask. Leftmost column shows the input mask. Columns 2,3,4 and 5 show the synthetic images generated using a single mask shown in the leftmost column.} 
    \label{fig:mask_multiple}
    \vspace{-2em}
    \end{figure}
    \begin{figure}[h]
    \begin{center}
      \includegraphics[width=0.85\textwidth]{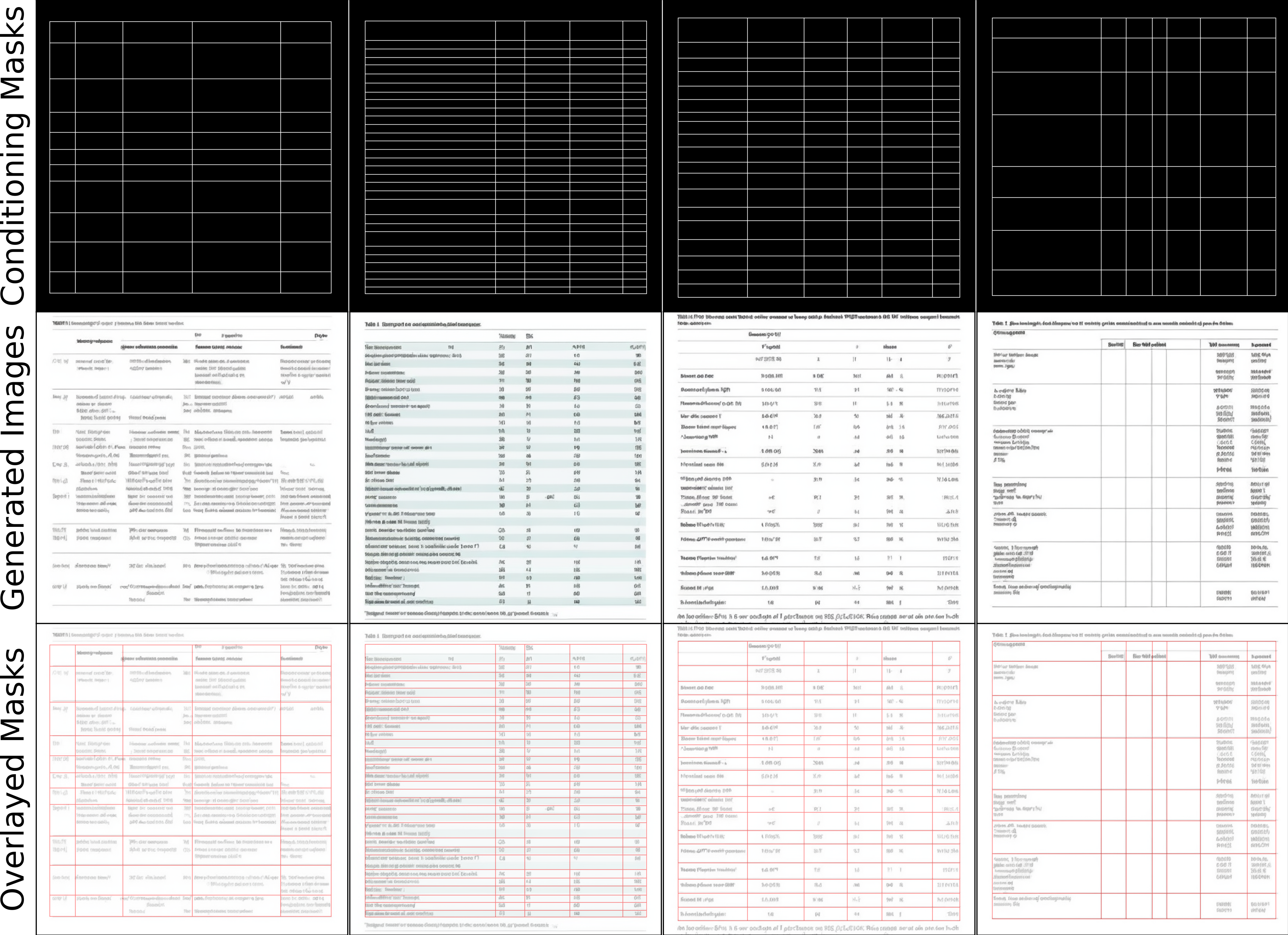}
    \end{center}
    \caption{Results of mask conditioned table image generations. Top row shows the input masks to the model. Second row shows the generated images. Third row shows the overlap of the mask on the generated image.}
    \label{fig:mask_different}
    \vspace{-1em}
    \end{figure}
	\begin{figure}[b!]
        \begin{center}
		  \includegraphics[width=0.85\textwidth]{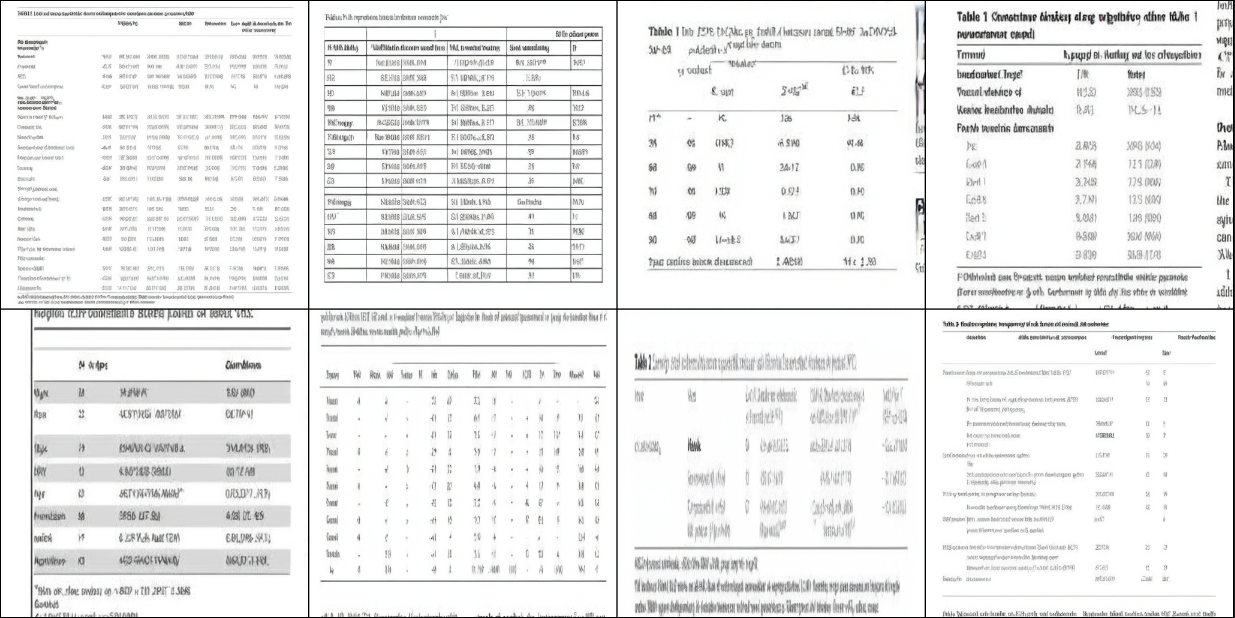}
        \end{center}
		\caption{Sample results of unconditional table image generations of size 512x512.} 
        \label{fig7}
        \vspace{-1em}
	\end{figure} 
     In this section, we present an extensive evaluation of our approach, from both quantitative and qualitative perspectives, and provide a thorough analysis of our findings.	
    \subsection{Qualitative Evaluation}

    In this section, we demonstrate the effectiveness of our proposed approach for table synthesis through a qualitative analysis. In Fig.~\ref{fig:mask_multiple}, we present a few sample images generated in a controlled manner utilizing the structure-conditioning masks. It is evident from the figure that for a single mask input, our approach is capable of generating a diverse variety of distinct-looking table images simply by using different initial noise seed values, all of which closely adhere to the target row/column structure provided by the input mask. 	
	Furthermore, it is worth noting that the masks used in this scenario were randomly generated, making them entirely out-of-distribution and unseen by the respective model during the training phase. This demonstrates that our proposed model was able to successfully learn appropriate relationships between the input structure and the target images, which allowed it to generalize effectively to out-of-distribution masks as well.

    In Fig.~\ref{fig:mask_different}, we present additional results with input conditioning masks overlaid on top of the generated images. It can be observed that our approach was effective in closely aligning the structure of the generated table with the input mask condition, making it suitable for synthesizing a diverse variety of new table images with pre-annotated structure. Note that the input conditioning masks were randomly generated in this case as well.	
	For comparison, we also illustrate the results of unconditional generation with $512\times 512$ resolution in Fig.~\ref{fig7}.  It can be observed that while the model is capable of producing sufficiently realistic tables through unconditional generation, there is no control over their structure.

	\begin{table}[t!]
		\centering
		\caption{FID scores achieved on our synthetic data.}\label{tab2}
			\setlength{\tabcolsep}{0.5em} %
			\begin{tabular}{cccc}
				\toprule
				Image Dimension &  Mask Conditioning & Generated Images & FID score\\
				\midrule
				256x256 &  \xmark & 250k & 9.40 \\
				512x512 &  \xmark & 250k & 8.38 \\
				256x256 &  \cmark & 250k & 8.22 \\
				512x512 &  \cmark & 250k &  7.81\\
				\bottomrule
			\end{tabular}
	\end{table}
	\begin{table}[b!]
		\centering
		\caption{mAP scores achieved using our synthetic data to train yolov5 for table structure recognition task. The model was trained using the synthetic data (512x512) obtained from mask-conditioned model shown in ~\ref{tab2}}\label{tab3}
		\setlength{\tabcolsep}{0.5em} %
		\begin{tabular}{cccc}
			\toprule
			Model &  Training Dataet & $mAP_{50}$ & $mAP_{75}$\\
			\midrule
			Faster R-CNN~\cite{fasterrcnn} & \pubtables{} & 0.815 & 0.785 \\
			DETR-100k~\cite{detr} & \pubtables{} & 0.963 & 0.932 \\
			DETR~\cite{detr} & \pubtables{} & 0.971 & 0.948 \\
			YOLOv5~\cite{yolo} & \pubtables{} & 0.970 & 0.944 \\
			YOLOv5~\cite{yolo} & Synthetic-250k (Ours) & \textbf{0.947} &  \textbf{0.918}\\
			\bottomrule
		\end{tabular}
	\end{table} 
	\begin{table}[t!]
		\centering
		\caption{Evaluation results of trained YOLOv5 model compared with competitors from ICDAR2013 competition. For simplicity only the results of top 3 competitors from the competition have been displayed.}\label{tab4}
		\resizebox{\textwidth}{!}{
			\setlength{\tabcolsep}{0.5em} %
			\begin{tabular}{ccccc}
				\toprule
				Model/Participant & Training Dataset & Recall & Precision & F1-measure\\
				\midrule
				FineReader~\cite{ICDAR-2013} & ICDAR2013~\cite{ICDAR-2013} & 0.8835 &  0.8710 & 0.8772 \\
				OmniPage~\cite{ICDAR-2013}  &  ICDAR2013~\cite{ICDAR-2013} & 0.8380 &  0.8460 & 0.8420 \\
				Nurminen~\cite{ICDAR-2013}  & ICDAR2013~\cite{ICDAR-2013} & 0.8078 &  0.8693 & 0.8374 \\
				YOLOv5~\cite{yolo} & ICDAR2013~\cite{ICDAR-2013} & 0.9322 & 0.9379 & 0.9350 \\
				YOLOv5~\cite{yolo} & Synthetic-1k (Ours) & 0.9014 & 0.9131 & 0.9072 \\
				YOLOv5~\cite{yolo} & ICDAR2013~\cite{ICDAR-2013} + Synthetic-1k (Ours) & \textbf{0.9510} & \textbf{0.9562} & \textbf{0.9536} \\
				\bottomrule
			\end{tabular}
		}
	\end{table} 
	\subsection{Quantitative Evaluation}
	We evaluate the generated synthetic data from our trained models in two different scenarios. Firstly, we compute the FID scores on the synthetic data, as shown in Table~\ref{tab2}. The FID scores are calculated separately for both conditional and unconditional table image synthesis, across dataset sizes of 250K.

    Table~\ref{tab2} clearly demonstrates that the quality of table images generated using conditioning techniques is superior to those generated unconditionally. The low FID scores for conditional table image generations, across both dimensions (256x256 and 512x512), compared to unconditional table image generations, suggest that our methodology enhances the overall structure of the synthetic images and produces coherent results.

    In the second scenario, we train an object detection model called YOLOv5 ~\cite{yolo} on our synthetic data for the task of table structure recognition. It is important to note that the masks we used to condition our synthetic data generation process serve as the ground-truth annotations for the generated data. After training the YOLOv5 ~\cite{yolo} model using our generated synthetic images and their corresponding annotations derived from their masks, we assess the performance of the trained model on the test set of the PubTables-1M dataset \cite{pubtables}, comprising 93,834 images. Similarly, we compare the performance of this model with that of YOLOv5 ~\cite{yolo} trained on the original training set of the PubTables-1M dataset, which includes 758,849 training images. The evaluation metric used here is mean average precision (mAP) using two different Intersection over Union (IoU) thresholds (50 and 75). The mAP values are provided in Table~\ref{tab3}, along with the performance of state-of-the-art models \cite{pubtables} on the same test set.

    We further assess the usability and viability of our data on the ICDAR2013 table competition dataset \cite{ICDAR-2013}. We evaluate the performance of three YOLOv5 ~\cite{yolo} models: one trained on the validation set of the ICDAR2013 competition dataset \cite{ICDAR-2013}, which consists of 100 annotated images for table structure, a second one trained on 1K synthetic table images generated using the masked images from the ICDAR2013 \cite{ICDAR-2013} competition validation set and finally a third model which is trained by augmenting the validation set of ICDAR2013 competition dataset \cite{ICDAR-2013} with the synthetic data we generated for the previous model. The masked images were generated using the same technique described in the data preparation section. All three models are evaluated on the test set of the ICDAR2013 competition table structure recognition dataset \cite{ICDAR-2013} which consists of 158 images. As observed from the results in Table~\ref{tab3}, the base model of yolov5 ~\cite{yolo} trained on the validation set performs extremely well whereas the model which is trained only on our 1k synthetic table images performs better than the participants of the ICDAR2013 competition participants \cite{ICDAR-2013} and finally the yolov5 ~\cite{yolo} model trained on validation plus our 1k synthetic table images outperforms all of the rest which further strengthens our proposed approach and signifies that the synthetic data generated using our methodology is extremely useful and viable.

    Despite achieving high mAP and F1 scores using models trained on our synthetic data, it should be noted that there remains a minor performance gap compared to models trained on real data. The results in the qualitative evaluation section will show that, even though we are able to obtain high-resolution (512x512) document table images, the text still appears smudged in the generated images. Further improvement can be achieved by training latent diffusion models with our methodology for more iterations than we did in Table~\ref{tab:exp_setup}. Overall, it is evident that our synthetic data consistently delivered strong performance across different scenarios, including the PubTables-1M and ICDAR2013 competition datasets.

    \section{Conclusion}
    This papers presents a novel method for conditioning table images through their table structure masks which allow latent diffusion models to better understand their underlying data distribution and allows us to generate new synthetic table images along with their table structure annotations. The synthetic data generated from our approach was tested for its viability using FID scores which were very low indicating that the quality of our generated synthetic data was really good. Furthermore, the synthetic data was also used to train an object detection model called yolov5 to identify table structure (rows/columns) in unseen images. This trained model was tested on the pubtables-1m testset and on the ICDAR2013 competition dataset, and it provided results which were closely aligned with state of the art models and much better in the case of ICDAR2013 competition participants. The results that we have achieved using conditioned LDMs are very promising but there are still some limitations which need to be discussed. For example, the quality of text inside the tables is not coherent and it can be improved further by training the model for 700-800K iterations. Furthermore, the conditioning process can also be modified to accept text input about number of rows and columns in the table image that needs to be generated which will allow more fine-grained control over the generation process.


\begin{thebibliography}{10}
	\providecommand{\url}[1]{\texttt{#1}}
	\providecommand{\urlprefix}{URL }
	\providecommand{\doi}[1]{https://doi.org/#1}
	
	\bibitem{d3pm}
	Austin, J., Johnson, D.D., Ho, J., Tarlow, D., van~den Berg, R.: Structured
	denoising diffusion models in discrete state-spaces (2023)
	
	\bibitem{detr}
	Carion, N., Massa, F., Synnaeve, G., Usunier, N., Kirillov, A., Zagoruyko, S.:
	End-to-end object detection with transformers. CoRR  \textbf{abs/2005.12872}
	(2020), \url{https://arxiv.org/abs/2005.12872}
	
	\bibitem{ICDAR-2017}
	Gao, L., Yi, X., Jiang, Z., Hao, L., Tang, Z.: Icdar2017 competition on page
	object detection. 2017 14th IAPR International Conference on Document
	Analysis and Recognition (ICDAR)  \textbf{01},  1417--1422 (2017),
	\url{https://api.semanticscholar.org/CorpusID:34640499}
	
	\bibitem{gans}
	Goodfellow, I., Pouget-Abadie, J., Mirza, M., Xu, B., Warde-Farley, D., Ozair,
	S., Courville, A., Bengio, Y.: Generative adversarial nets. In: Ghahramani,
	Z., Welling, M., Cortes, C., Lawrence, N., Weinberger, K. (eds.) Advances in
	Neural Information Processing Systems. vol.~27. Curran Associates, Inc.
	(2014),
	\url{https://proceedings.neurips.cc/paper_files/paper/2014/file/5ca3e9b122f61f8f06494c97b1afccf3-Paper.pdf}
	
	\bibitem{ICDAR-2013}
	Göbel, M., Hassan, T., Oro, E., Orsi, G.: Icdar 2013 table competition. In:
	2013 12th International Conference on Document Analysis and Recognition. pp.
	1449--1453 (2013). \doi{10.1109/ICDAR.2013.292}
	
	\bibitem{diff-layout-2}
	He, L., Lu, Y., Corring, J., Florencio, D., Zhang, C.: Diffusion-based document
	layout generation. In: Fink, G.A., Jain, R., Kise, K., Zanibbi, R. (eds.)
	Document Analysis and Recognition - ICDAR 2023. pp. 361--378. Springer Nature
	Switzerland, Cham (2023)
	
	\bibitem{ddpm}
	Ho, J., Jain, A., Abbeel, P.: Denoising diffusion probabilistic models. In:
	Proceedings of the 34th International Conference on Neural Information
	Processing Systems. NIPS'20, Curran Associates Inc., Red Hook, NY, USA (2020)
	
	\bibitem{diff-layout-1}
	Inoue, N., Kikuchi, K., Simo-Serra, E., Otani, M., Yamaguchi, K.: Layoutdm:
	Discrete diffusion model for controllable layout generation. In: Proceedings
	of the IEEE/CVF Conference on Computer Vision and Pattern Recognition (CVPR).
	pp. 10167--10176 (June 2023)
	
	\bibitem{stylegan}
	Karras, T., Aittala, M., Laine, S., H\"ark\"onen, E., Hellsten, J., Lehtinen,
	J., Aila, T.: Alias-free generative adversarial networks. In: Proc. NeurIPS
	(2021)
	
	\bibitem{vaes}
	Kingma, D.P., Welling, M.: {Auto-Encoding Variational Bayes}. In: 2nd
	International Conference on Learning Representations, {ICLR} 2014, Banff, AB,
	Canada, April 14-16, 2014, Conference Track Proceedings (2014)
	
	\bibitem{tablebank}
	Li, M., Cui, L., Huang, S., Wei, F., Zhou, M., Li, Z.: {T}able{B}ank: Table
	benchmark for image-based table detection and recognition. In: Calzolari, N.,
	B{\'e}chet, F., Blache, P., Choukri, K., Cieri, C., Declerck, T., Goggi, S.,
	Isahara, H., Maegaard, B., Mariani, J., Mazo, H., Moreno, A., Odijk, J.,
	Piperidis, S. (eds.) Proceedings of the Twelfth Language Resources and
	Evaluation Conference. pp. 1918--1925. European Language Resources
	Association, Marseille, France (May 2020),
	\url{https://aclanthology.org/2020.lrec-1.236}
	
	\bibitem{tsrformer}
	Lin, W., Sun, Z., Ma, C., Li, M., Wang, J., Sun, L., Huo, Q.: Tsrformer: Table
	structure recognition with transformers (2022)
	
	\bibitem{luhman2020diffusion}
	Luhman, T., Luhman, E.: Diffusion models for handwriting generation (2020)
	
	\bibitem{tableformer}
	Nassar, A., Livathinos, N., Lysak, M., Staar, P.: Tableformer: Table structure
	understanding with transformers. In: 2022 IEEE/CVF Conference on Computer
	Vision and Pattern Recognition (CVPR). pp. 4604--4613 (2022).
	\doi{10.1109/CVPR52688.2022.00457}
	
	\bibitem{diffusion-based-text-gen}
	Nikolaidou, K., Retsinas, G., Christlein, V., Seuret, M., Sfikas, G., Smith,
	E.B., Mokayed, H., Liwicki, M.: Wordstylist: Styled verbatim handwritten text
	generation with latent diffusion models (2023)
	
	\bibitem{pixelcnn}
	van~den Oord, A., Kalchbrenner, N., Vinyals, O., Espeholt, L., Graves, A.,
	Kavukcuoglu, K.: Conditional image generation with pixelcnn decoders (2016)
	
	\bibitem{vqvae}
	van~den Oord, A., Vinyals, O., kavukcuoglu, k.: Neural discrete representation
	learning. In: Guyon, I., Luxburg, U.V., Bengio, S., Wallach, H., Fergus, R.,
	Vishwanathan, S., Garnett, R. (eds.) Advances in Neural Information
	Processing Systems. vol.~30. Curran Associates, Inc. (2017),
	\url{https://proceedings.neurips.cc/paper_files/paper/2017/file/7a98af17e63a0ac09ce2e96d03992fbc-Paper.pdf}
	
	\bibitem{diffTransformer}
	Peebles, W., Xie, S.: Scalable diffusion models with transformers (2023)
	
	\bibitem{ShabbyPages2023}
	Project, T.A.: Shabbypages: A reproducible document denoising and binarization
	dataset (2023), \url{https://github.com/sparkfish/shabby-pages}
	
	\bibitem{dcgan}
	Radford, A., Metz, L., Chintala, S.: Unsupervised representation learning with
	deep convolutional generative adversarial networks (2015),
	\url{http://arxiv.org/abs/1511.06434}, cite arxiv:1511.06434Comment: Under
	review as a conference paper at ICLR 2016
	
	\bibitem{dalle2}
	Ramesh, A., Dhariwal, P., Nichol, A., Chu, C., Chen, M.: Hierarchical
	text-conditional image generation with clip latents (2022)
	
	\bibitem{dalle}
	Ramesh, A., Pavlov, M., Goh, G., Gray, S., Voss, C., Radford, A., Chen, M.,
	Sutskever, I.: Zero-shot text-to-image generation. In: Meila, M., Zhang, T.
	(eds.) Proceedings of the 38th International Conference on Machine Learning.
	Proceedings of Machine Learning Research, vol.~139, pp. 8821--8831. PMLR
	(18--24 Jul 2021), \url{https://proceedings.mlr.press/v139/ramesh21a.html}
	
	\bibitem{yolo}
	Redmon, J., Divvala, S., Girshick, R., Farhadi, A.: You only look once:
	Unified, real-time object detection (2015),
	\url{http://arxiv.org/abs/1506.02640}, cite arxiv:1506.02640
	
	\bibitem{fasterrcnn}
	Ren, S., He, K., Girshick, R.B., Sun, J.: Faster {R-CNN:} towards real-time
	object detection with region proposal networks. CoRR  \textbf{abs/1506.01497}
	(2015), \url{http://arxiv.org/abs/1506.01497}
	
	\bibitem{ldms}
	Rombach, R., Blattmann, A., Lorenz, D., Esser, P., Ommer, B.: High-resolution
	image synthesis with latent diffusion models. In: Proceedings of the IEEE/CVF
	Conference on Computer Vision and Pattern Recognition (CVPR). pp.
	10684--10695 (June 2022)
	
	\bibitem{saharia2022palette}
	Saharia, C., Chan, W., Chang, H., Lee, C.A., Ho, J., Salimans, T., Fleet, D.J.,
	Norouzi, M.: Palette: Image-to-image diffusion models (2022),
	\url{https://openreview.net/forum?id=FPGs276lUeq}
	
	\bibitem{imagen}
	Saharia, C., Chan, W., Saxena, S., Li, L., Whang, J., Denton, E.L.,
	Ghasemipour, S.K.S., Ayan, B.K., Mahdavi, S.S., Lopes, R.G., Salimans, T.,
	Ho, J., Fleet, D.J., Norouzi, M.: Photorealistic text-to-image diffusion
	models with deep language understanding. ArXiv  \textbf{abs/2205.11487}
	(2022), \url{https://api.semanticscholar.org/CorpusID:248986576}
	
	\bibitem{saifullah-robustness}
	Saifullah, Siddiqui, S.A., Agne, S., Dengel, A., Ahmed, S.: Are deep models
	robust against real distortions? a case study on document image
	classification. In: 2022 26th International Conference on Pattern Recognition
	(ICPR). pp. 1628--1635 (2022). \doi{10.1109/ICPR56361.2022.9956167}
	
	\bibitem{diff-binarization}
	Saifullah, S., Agne, S., Dengel, A., Ahmed, S.: Coldbin: Cold diffusion
	for document image binarization. In: Fink, G.A., Jain, R., Kise, K.,
	Zanibbi, R. (eds.) Document Analysis and Recognition - ICDAR 2023. pp.
	207--226. Springer Nature Switzerland, Cham (2023)
	
	\bibitem{schuhmann2022laionb}
	Schuhmann, C., Beaumont, R., Vencu, R., Gordon, C.W., Wightman, R., Cherti, M.,
	Coombes, T., Katta, A., Mullis, C., Wortsman, M., Schramowski, P., Kundurthy,
	S.R., Crowson, K., Schmidt, L., Kaczmarczyk, R., Jitsev, J.: {LAION}-5b: An
	open large-scale dataset for training next generation image-text models. In:
	Thirty-sixth Conference on Neural Information Processing Systems Datasets and
	Benchmarks Track (2022), \url{https://openreview.net/forum?id=M3Y74vmsMcY}
	
	\bibitem{layout-shen2021}
	Shen, Z., Zhang, R., Dell, M., Lee, B.C.G., Carlson, J., Li, W.: Layoutparser:
	A unified toolkit for deep learning based document image analysis. In:
	Llad{\'o}s, J., Lopresti, D., Uchida, S. (eds.) Document Analysis and
	Recognition -- ICDAR 2021. pp. 131--146. Springer International Publishing,
	Cham (2021)
	
	\bibitem{pubtables}
	Smock, B., Pesala, R., Abraham, R.: Pubtables-1m: Towards comprehensive table
	extraction from unstructured documents (2021)
	
	\bibitem{song2021denoising}
	Song, J., Meng, C., Ermon, S.: Denoising diffusion implicit models. In:
	International Conference on Learning Representations (2021),
	\url{https://openreview.net/forum?id=St1giarCHLP}
	
	\bibitem{diff-image-gen}
	Tanveer, N., Ul-Hasan, A., Shafait, F.: Diffusion models for document image
	generation. In: Fink, G.A., Jain, R., Kise, K., Zanibbi, R. (eds.) Document
	Analysis and Recognition - ICDAR 2023. pp. 438--453. Springer Nature
	Switzerland, Cham (2023)
	
	\bibitem{thanhtung2020catastrophic}
	Thanh-Tung, H., Tran, T.: Catastrophic forgetting and mode collapse in gans.
	2020 International Joint Conference on Neural Networks (IJCNN) pp. 1--10
	(2020), \url{https://api.semanticscholar.org/CorpusID:221659882}
	
	\bibitem{transformers}
	Vaswani, A., Shazeer, N., Parmar, N., Uszkoreit, J., Jones, L., Gomez, A.N.,
	Kaiser, L.u., Polosukhin, I.: Attention is all you need. In: Guyon, I.,
	Luxburg, U.V., Bengio, S., Wallach, H., Fergus, R., Vishwanathan, S.,
	Garnett, R. (eds.) Advances in Neural Information Processing Systems.
	vol.~30. Curran Associates, Inc. (2017),
	\url{https://proceedings.neurips.cc/paper_files/paper/2017/file/3f5ee243547dee91fbd053c1c4a845aa-Paper.pdf}
	
	\bibitem{tables-survey}
	Xiao, B., Simsek, M., Kantarci, B., Alkheir, A.A.: Revisiting table detection
	datasets for visually rich documents (2023)
	
	\bibitem{diff-enhancement}
	Yang, Z., Liu, B., Xiong, Y., Yi, L., Wu, G., Tang, X., Liu, Z., Zhou, J.,
	Zhang, X.: Docdiff: Document enhancement via residual diffusion models (2023)
	
	\bibitem{diffusion-based-text-gen-2}
	Zhang, L., Rao, A., Agrawala, M.: Adding conditional control to text-to-image
	diffusion models (2023)
	
	\bibitem{fintabnet}
	Zheng, X., Burdick, D., Popa, L., Zhong, X., Wang, N.X.R.: Global table
	extractor (gte): A framework for joint table identification and cell
	structure recognition using visual context. In: Proceedings of the IEEE/CVF
	Winter Conference on Applications of Computer Vision (WACV). pp. 697--706
	(January 2021)
	
	\bibitem{publaynet}
	Zhong, X., Tang, J., Yepes, A.J.: Publaynet: Largest dataset ever for document
	layout analysis. In: 2019 International Conference on Document Analysis and
	Recognition (ICDAR). pp. 1015--1022. IEEE Computer Society, Los Alamitos, CA,
	USA (sep 2019). \doi{10.1109/ICDAR.2019.00166},
	\url{https://doi.ieeecomputersociety.org/10.1109/ICDAR.2019.00166}
	
	\bibitem{pubtabnet}
	Zhong, X., ShafieiBavani, E., Yepes, A.J.: Image-based table recognition: data,
	model, and evaluation. arXiv preprint arXiv:1911.10683  (2019)
	
	\bibitem{scitsr}
	Zhong, X., ShafieiBavani, E., Yepes, A.J.: Image-based table recognition: data,
	model, and evaluation. arXiv preprint arXiv:1911.10683  (2019)
	
\end{thebibliography}
\end{document}